\documentclass{article}
\usepackage{spconf,amsmath,graphicx}

\usepackage{caption} 
\usepackage[normalem]{ulem}
\usepackage{arydshln}
\usepackage{caption} 

\usepackage{cleveref}

\usepackage{amssymb}
\usepackage{amsmath}

\DeclareMathOperator*{\argmin}{arg\,min}

\title{HYPERNETWORK-BASED ADAPTIVE IMAGE RESTORATION}
\name{Shai Aharon~~~~~ Gil Ben-Artzi}
\address{Department of Computer Science\\
	Ariel University, Israel}
\begin{document}
%\ninept
%
\maketitle
\begin{abstract}
Adaptive image restoration models can restore images with different degradation levels at inference time without the need to retrain the model. We present an approach that is highly accurate and allows a significant reduction in the number of parameters. In contrast to existing methods, our approach can restore images using a single fixed-size model, regardless of the number of degradation levels. On popular datasets, our approach yields state-of-the-art results in terms of size and accuracy for a variety of image restoration tasks, including denoising, deJPEG, and super-resolution.
\end{abstract}
\section{Introduction}
\label{sec:intro}
The common approach in deep learning for image restoration tasks is to train the model in a supervised manner, optimizing the model for only a single degradation level. The exact degradation level of the degraded image is not known a priori, and treating all degradation levels the same lowers restoration quality. Recently, adaptive image restoration methods have gained popularity as an alternative. They enable image- and user-specific adaptation for all degradation levels with a single model without the need for retraining or deploying multiple models. At runtime, the user can adjust the restoration effect in order to generate a variety of output images and select one according to his preferences.

Various approaches have been presented for adaptive models. State-of-the-art approaches~\cite{He_2019_CVPR,dnInterpolation,dynamicnet,wang2019cfsnet} reduce the needed number of parameters with respect to multiple independent models by training the network to restore two degradation levels that span the desired range and interpolate the weights for any other degradation level within the range. However, to support a wide range of levels it is necessary to use several models or significantly increase the size of the basic model ~\cite{cresmd}.

%To ensure an efficient representation at inference time, we constrain the kernels corresponding to the different levels to be identical up to a scaling factor and learn only the weights' basis vectors. 

%We present a novel approach in this work that allows a standard single restoration model to achieve optimal accuracy across a wide range of degradation levels without adding any additional parameters. In this paper, we introduce a hypernetwork that can learn to generate filter weights for an image restoration network based on an input parameter that specifies the required restoration level. As part of the training process, our model is trained with multiple main networks to simultaneously restore images with a variety of degradation levels

In this work, we present an approach that allows a standard single restoration model to achieve very high accuracy across a wide range of degradation levels without having to add any extra parameters. We introduce a hypernetwork that learns to generate the filter weights of an image restoration network conditionally based on the required restoration level given as an input parameter. As part of the training process, our hypernetwork is optimized with multiple main networks to simultaneously restore images with a variety of degradation levels. Based on the input degradation level, our hypernetwork generates a single network with the most accurate filter weights at runtime.

{\bf Contribution.} Our architecture can restore images of different degrees of degradation with 26\%-56\% of the parameters and higher accuracy than existing adaptive image restoration methods.

\section{Prior Work}

Recently, there has been a growing interest in constructing networks that can be continuously tuned at inference time. These can broadly be categorized into two categories, models which allow tuning different objectives at inference time and models which allow different restoration levels of the same objective, where our approach falls into the latter category. The typical approach is to train two related networks on different objectives and apply interpolation between their weights. The networks can be either the same or with additional tuning blocks. Dynamic-Net \cite{dynamicnet} adds specialized blocks directly after the convolution layers, which are optimized during the training for the additional objective. 
CFSNet \cite{wang2019cfsnet} used branches, each based on a different objective. 
AdaFM \cite{He_2019_CVPR} added modulation filters after each convolution layer. DNI \cite{dnInterpolation} train the same network architecture on different objective and interpolates all the parameters. 
Son \cite{lee2020smoother} extended the approach of \cite{He_2019_CVPR} with an FTN module allowing better non-linear interpolation. \cite{meta_sr} generates kernels for the super-resolution task. They employ an off-the-shelf SR network \cite{rdn} as a backbone which is $\times 10$ bigger than our network, and their proposed solution is applicable only to the super-resolution task. ~\cite{jiang2021png} proposed solving the image restoration task as a multi-task problem. Their network, however, is specialized for restoring only a limited number of degradation levels, each of which must be trained individually. In our approach, the model is trained on a small set of degradation levels and can continuously restore any other level inside and even outside this range.

Learning to learn, or meta learner, uses meta networks to generate weights for the main network for various tasks \cite{liu2020metadistiller}.  Hypernetwork, introduced in \cite{ha2016hypernetworks}, uses a small network with a reduced number of parameters to generate the weights for a larger target network. It often uses weight sharing across layers, while providing accurate results. \cite{fan2018decouple} presented an image restoration hypernetwork with a single main network. In our approach, we employ a hypernetwork to generate the weights of kernels in multiple target networks simultaneously.

\section{Method}

\subsection{Formulation}

Our model consists of a hypernetwork $h$ and main restoration networks $n_i$. The weights of our hypernetwork, $\theta^h$, are learned during the training process and fixed during inference time. The input to hypernetwork $h$ is a degradation level $c_i \in \mathbb{R}$ and the output is $\theta^{n_i}$, the kernels' weights for the main restoration networks $n_i$, $\theta^{n_i}=h(c_i;\theta^{h})$. The input for each main network $n_i$ is a degraded image ${\bf I}^{c_i} \in \mathbb{R}^{3 \times H \times W}$ with a degradation level $c_i$, $H,W$ are the height and width of the image and the output is the restored image with the same dimensions, $n_i({\bf I}^{c_i};\theta^{n_i})$. The goal is to learn $\theta^h$ from ${\bf I}^{c_i},c_i$ so that $h$ can generate the optimal weights for the corresponding restoration network $\theta^{n_i}$ in order to restore the degraded image. The optimization problem associated with our model is formulated as:
\begin{gather}
    \argmin_{\theta^{h}}{\sum_{i=1}^k \mathbb{E}_{{\bf I}^{c_i},{\bf I}}\big[\mathcal{L}\big ( n_i({\bf I}^{c_i};h(c_i;\theta^{h})),{\bf I} \big) \big]}
\end{gather}

where $k$ is the number of main networks, $\mathcal{L}: \mathbb{R}^{3 \times H \times W} \times \mathbb{R}^{3 \times H \times W} \rightarrow \mathbb{R}_{+} \cup \{0\}$ is our restoration loss for each main network. We train our hypernetwork, $h$, by simultaneously generating the weights for all the $k$ main networks (and degradation levels), and optimizing them together. We demonstrate that this enables our hypernetwork to generate optimal weights for all other degradation levels within the continuous range that are not included in the $k$ levels.

\subsection{The Network}

{\bf Parametrization of Main Network.} The hypernetwork $h$ consists of $l$ meta blocks, where $l$ is the number of kernels in the main network's residual blocks. Each meta block is a fully connected layer constructed out of weights and biases, ${\bf w}^j, {\bf b}^j \in \mathbb{R}^{ (C_{out} \times C_{in} \times K \times K)  \times 1 }$, where $C_{in}$ and $C_{out}$ are the number of input and output channels, respectively, and $K \times K$ is the kernel's size. The $j^{th}$ kernel of main network $n_i$ is ${\bf k}^j_i$, of the same dimensions as the meta block. The parameters of the hypernetwork and each main network are $\theta^h=\{({\bf w}^j,{\bf b}^j)\}_{j=1}^l$, $\theta^{n_i}=\{({\bf k}^j_i,
\}_{j=1}^l$.

We parameterize the kernels of the main network as a linear combination of our hypernetwork's weights and biases with the degradation level as a scaling parameter. Thus, our hypernetwork $h$ generates the weights of the $j^{th}$ kernel of main network $n_i$ for degradation level $c_i$ by: 
\begin{equation}
   {\bf k}^j_i = c_i {\bf w}^j + {\bf b}^j. 
    \label{eq:kernel}
\end{equation}

This allows a highly efficient representation as we only need to store the weights and biases of the hypernetwork in order to generate the corresponding kernels for each possible restoration network given the degradation level. We demonstrate that, despite the compact representation, our model is highly accurate. Note that, unlike hypernetworks \cite{ha2016hypernetworks}, our method assigns each meta block to generate weights for a specific main network layer with one common input scalar. Due to the bias term, the output convolutional kernels are not identical throughout the various main networks up to the input scalar.
%~\Cref{fig:hyperblock} presents the generation of the main network's kernels by the meta blocks. 

\begin{figure}[tb]
    \begin{center}
        \includegraphics[width=0.95\linewidth]{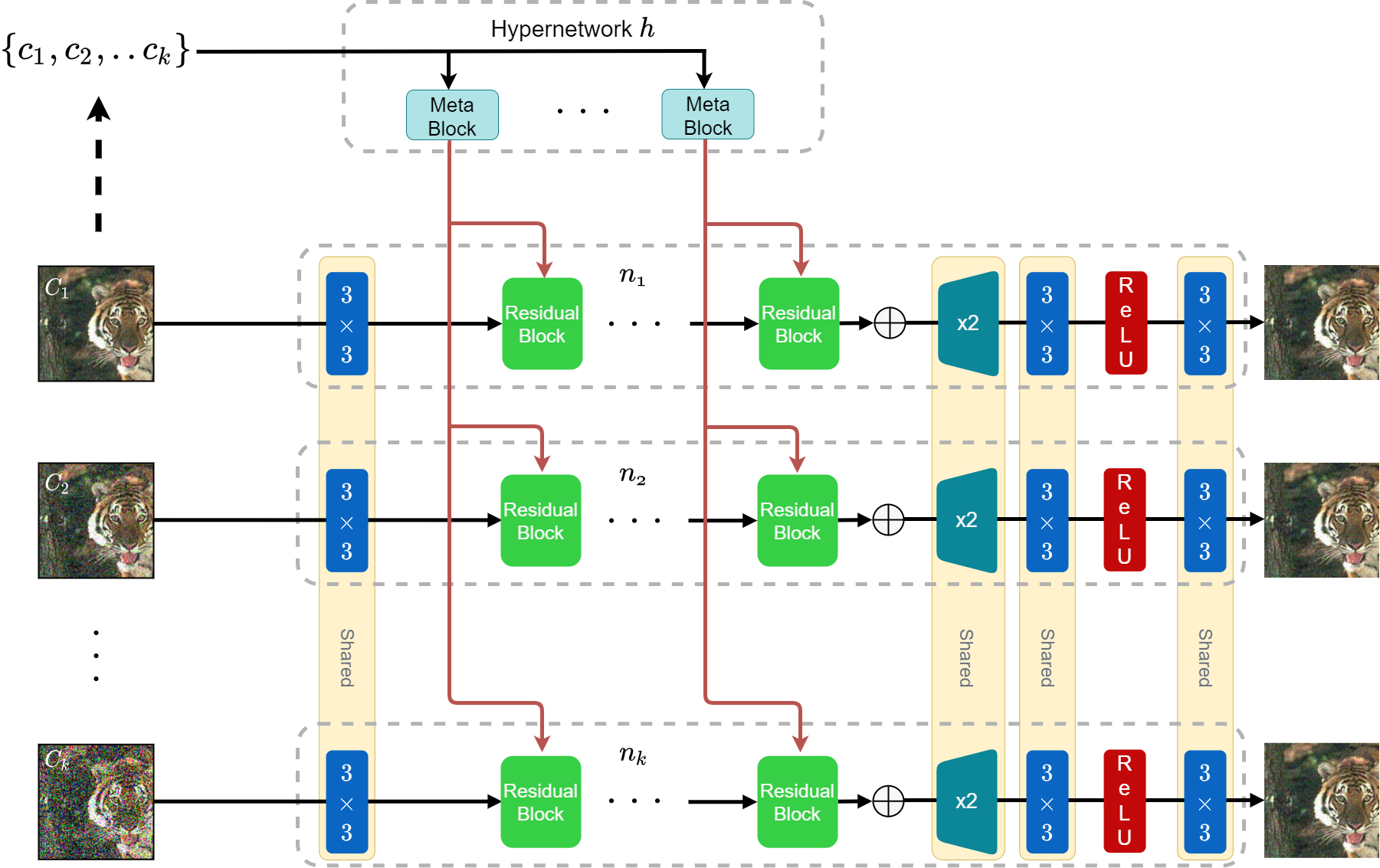}
    \end{center}
   \caption{The training framework. $k$ main networks are generated by the hypernetwork using the same kernels. Each is fed with a corresponding degraded image, and the losses are summed and back-propagated to the shared and hypernetwork weights. At runtime, a single main network is generated based on a single input parameter.}
\label{fig:train}
\end{figure}

{\bf Training.} During training, the hypernetwork generates multiple main networks, each main network $n_i$ is optimized to restore a degraded image with a corresponding degradation level $c_i$. The number of main networks ($k$) is fixed during the training process. The main network is a standard image restoration network \cite{arcnn}. It consists of a downsampling layer using convolution with a stride of 2, 16 residual \cite{he2016deep} blocks and upsampling layers using pixelshuffle \cite{shi2016real} and a skip-connection over the residual blocks. The weights of each main network are the weights of the residual block's kernels generated by the hypernetwork (Fig. \ref{fig:train}, green background) and the weights of the head and tail of the network which are shared among all the main networks (Fig. \ref{fig:train}, yellow background). 

Each image in the training set $\mathcal{D} = \{ {\bf I_1}, {\bf I_2}, \ldots, {\bf I_n}   \}$ is degraded with $k$ degradation levels $\{ c_1, c_2, \ldots, c_k \}$ and fed into the corresponding main network $\{ n_1, n_2, \ldots, n_k \}$. Each main network is generated according to the degradation level $c_i$ and meta blocks. Our goal is to optimize the overall restoration accuracy under the different degradation levels. Therefore, no degradation level is privileged and our total loss is the unweighted sum of individual $\mathcal{L}_1$ losses. Since the aforementioned weight generation operations are completely differentiable, the parameters in our hypernetwork $h$ are optimized simultaneously following the chain rule. The $L1$ loss is used throughout all the experiments. The training process is illustrated in Fig. \ref{fig:train}. 

{\bf Inference.} Given a degraded image and an input degradation level $c_i$, we employ the learned weights of the hypernetwork $\theta^{h}$ to generate the weights of a restoration network $\theta^{n_i}$. Each meta block generates the weights according to Eq.~\ref{eq:kernel}. A simple user interface enables the user to interact with the system in real-time, selecting the input value and, as a result, the desired restoration outcome. The restoration network generation is efficient due to the multiplication of the same single scalar for all the residual blocks' kernels of the main network.

\begin{table}[tb]
	\centering
	\footnotesize
	\caption{Results for DeJPEG artifacts removal task.}
	\begin{tabular}{lcccccc}

& \multicolumn{6}{c}{$PSNR$}\\
        \cline{2-7}
        & \uline{10} & \uline{30} & \uline{50} & \uline{70} & \uline{80} & \uline{Mean}\\
		Baseline & 28.82 &  32.57 & 34.40 & 36.40 & 38.14 & 34.06\\ \cdashline{2-6} 
%		CResMD & 26.97 & 30.6 & 32.43 & 34.46 & 36.30 & \textbf{32.15}\\ \cdashline{2-6}
		Ours   & 28.81 & 32.56 & 34.39 & 36.38 & 38.09 & 34.04 \\
    \end{tabular}
	\begin{tabular}{lccccc}
		& \multicolumn{5}{c}{$SSIM$}\\
        \cline{2-6}
        & \uline{10} & \uline{30} & \uline{50} & \uline{70} & \uline{80}\\

		Baseline &  0.82  & 0.91 & 0.94  & 0.96 & 0.97\\\cdashline{2-6} 
		%CResMD &    0.80 & 0.92 & \textbf{0.94} & 0.95 & 0.96\\ \cdashline{2-6} 
		Ours &      0.82 & 0.91& 0.94 & 0.96 & 0.97\\ 
    \end{tabular}
	\label{table:dejpeg_full}
\end{table}

\begin{table}[tb]
	\centering
		\footnotesize
		\caption{Results for image denoising. }
		\begin{tabular}{lccccccc}
             & \multicolumn{6}{c}{$PSNR$}\\
             \cline{2-7} & \uline{5}&\uline{25}&\uline{45}&\uline{65}&\uline{90}& \uline{Mean}\\ 
			
			Baseline & 40.48&31.42&28.64&27.06&25.73&30.66\\ \cdashline{2-6} 
			
%			CResMD & 40.15&31.30&28.51&26.89&25.26\\ \cdashline{2-6} 
			Ours & 40.39&31.40&28.51&27.06&25.73 & 30.61\\ 
			
		\end{tabular}
	    \begin{tabular}{lccccc}
	
             & \multicolumn{5}{c}{$SSIM$}\\
             \cline{2-6}
			& \uline{5}&\uline{25}&\uline{45}&\uline{65}&\uline{85}\\
			Baseline &0.98 &0.89&0.81&0.76&0.71\\ \cdashline{2-6}
%            CResMD & 0.84&0.82&0.78&0.73&0.68\\ \cdashline{2-6}
			Ours & 0.98&0.89&0.81&0.76&0.71

		\end{tabular}
		\label{table:noise_full}
\end{table}

\begin{table}[tb]
	\centering
	\footnotesize
	\caption{Results for super-resolution task.}
	
	\begin{tabular}{lcccccc}
		& \multicolumn{6}{c}{$PSNR$}\\
        \cline{2-7}
		& \uline{2} & \uline{3} & \uline{4} & \uline{5} & \uline{6} & \uline{Mean}\\
		
		Baseline & 36.95 & 29.86 & 29.54  & 25.67  & 25.06 & 29.41\\ \cdashline{2-6} 
%		CResMD & 32.49 & 26.80 &25.93 & 23.13 & 22.47 & \textbf{26.16} \\ \cdashline{2-6}
		Ours &   36.71& 29.77 & 29.48 & 25.63 & 24.92 &29.30 \\
		\end{tabular}
	\begin{tabular}{lccccc}
		& \multicolumn{5}{c}{$SSIM$}\\
        \cline{2-6}
		& \uline{2} & \uline{3} & \uline{4} & \uline{5} & \uline{6}\\

		Baseline &  0.94 & 0.84 & 0.84 & 0.74 & 0.71\\ \cdashline{2-6}
%		CResMD &    0.92&0.82&0.78&0.69&0.65\\ \cdashline{2-6} 
		Ours &      0.94 & 0.84 & 0.83 & 0.74 & 0.71 \\
		\end{tabular}

	\label{table:sr_full}
\end{table}

\section{Experiments} 

We demonstrate that our approach with a single network is equivalent to deploying multiple networks (5-11), each is optimized for a different degradation level. We also achieve the same or higher accuracy with a significant reduction in the number of parameters with respect to state-of-the-art adaptive models.

We evaluate our approach with the following tasks - denoising, DeJPEG and super-resolution, based on popular benchmarks. The DIV2K\cite{Agustsson_2017_CVPR_Workshops} dataset was used to train the models for all tasks. For evaluation, we use the CBSD68 dataset \cite{martin2001database} for denoising, the LIVE1 \cite{LIVE1} for DeJPEG and the Set5 \cite{set5} for super-resolution.

\subsection{Comparison with optimal accuracy}

We deploy five (super-resolution), eight (DeJPEG) and eleven (denoising) dedicated restoration models, each specifically trained to restore a single degradation level. The basic restoration network, both ours and for each of the dedicated models, includes 16 residual blocks and is based on ~\cite{arcnn}. 

For DeJPEG, we evaluate our model at eight different compression levels. Table~\ref{table:dejpeg_full} shows our results with respect to optimal accuracy obtained by training the independent models to restore each compression level. In our approach, we achieve very high PSNR and SSIM, with a negligible PSNR distance from optimal accuracy. For denoising, we evaluate our model with respect to all noise levels from $5$ to $110$ with intervals of $5$. Table~\ref{table:noise_full} shows our results for the denoising task. Our model achieves restoration accuracy equivalent to that of eleven dedicated models using only a single network. For super-resolution, we evaluate our model using five upscaling factors, \(\times 2, \times 3,\times 4, \times 5, \times 6\). Table~\ref{table:sr_full} shows our results. Similar to the other tasks, our method achieves state-of-the-art accuracy with only one basic restoration network.

\subsection{Comparison with small size adaptive models}

We compare our accuracy with respect to existing state-of-the-art smaller adaptive models, AdaFM \cite{He_2019_CVPR} and CFSNet \cite{wang2019cfsnet}. Despite not being optimal, these models achieve very high accuracy with a small number of parameters with respect to multiple networks.

\begin{figure}[tb]
	\begin{center}
		\includegraphics[width=0.95\linewidth]{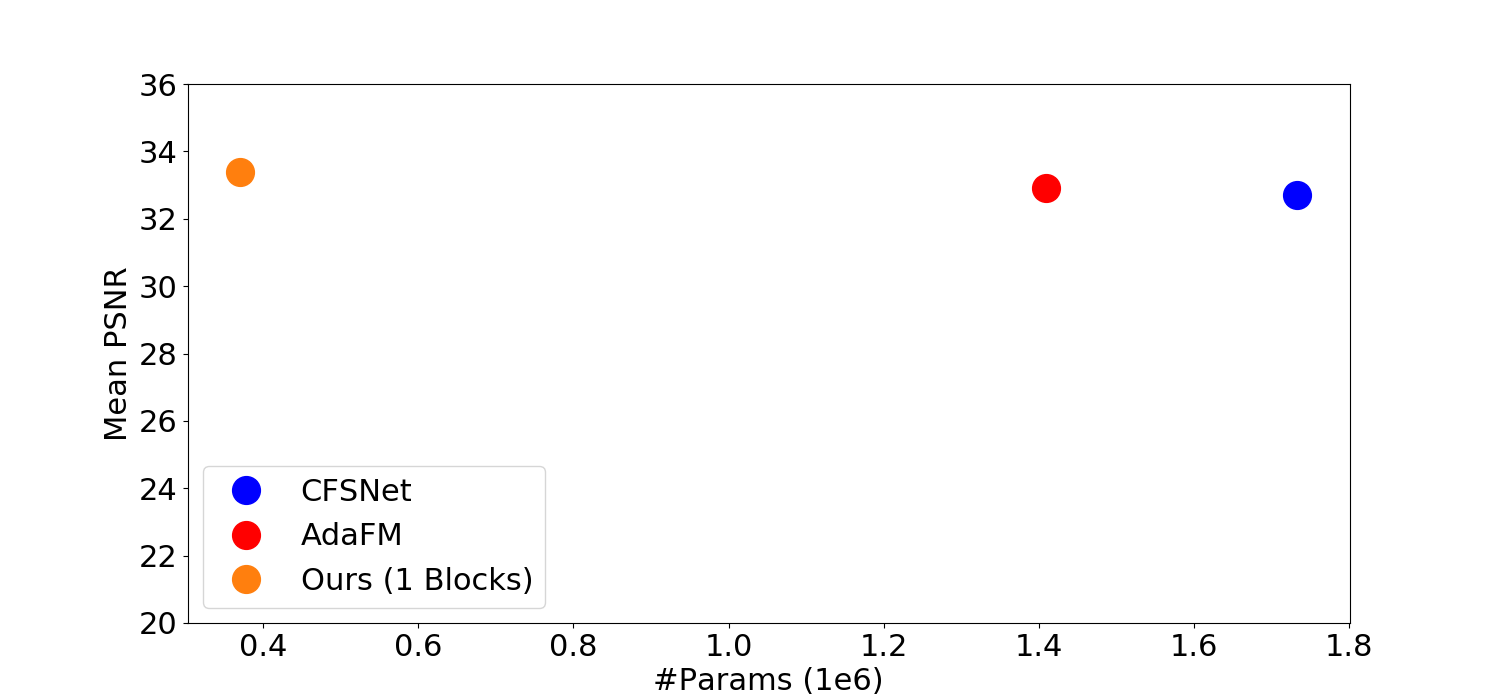}
	\end{center}
	\caption{Results for DeJPEG}
	\label{fig:weight_jpeg}
\end{figure}

\begin{figure}[tb]
	\begin{center}
		\includegraphics[width=0.95\linewidth]{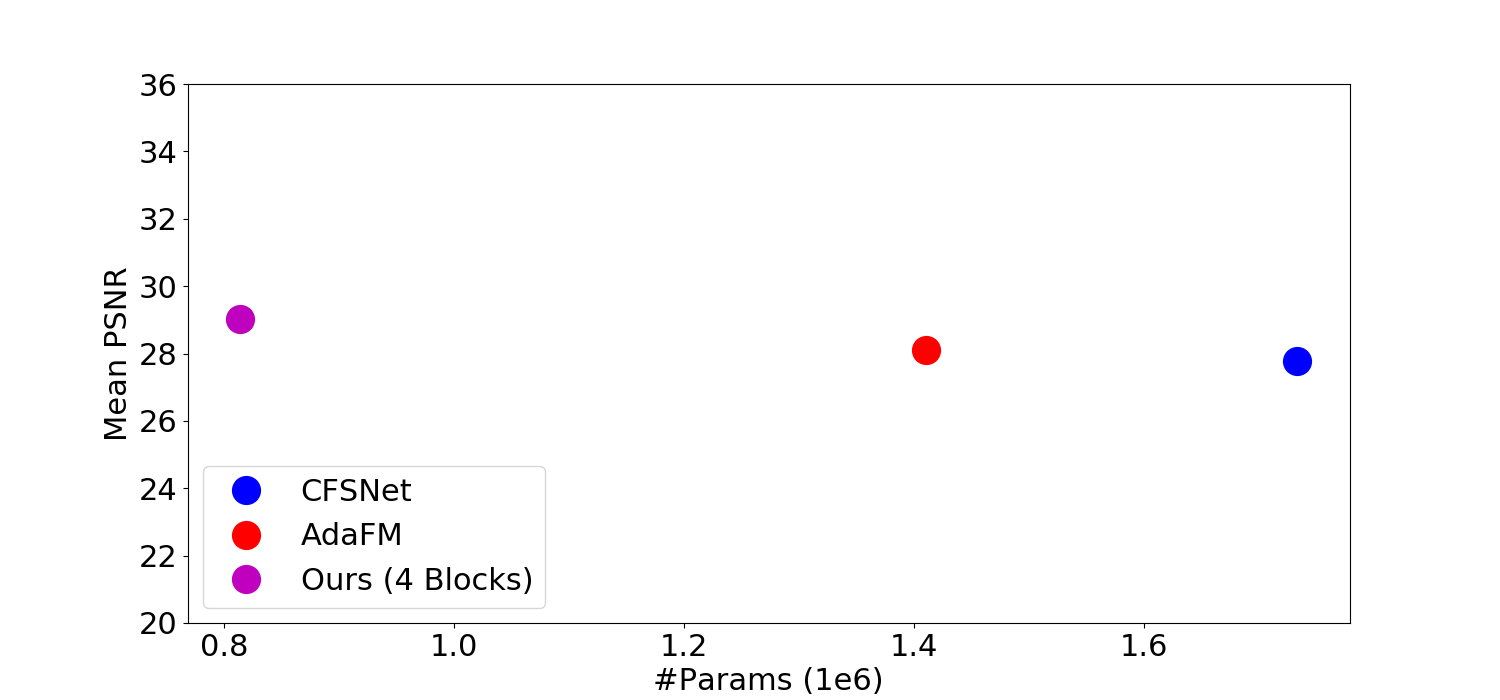}
	\end{center}
\caption{Results for denoising}
	\label{fig:weight_noise}
\end{figure}

\begin{figure}[tb]
	\begin{center}
		\includegraphics[width=0.95\linewidth]{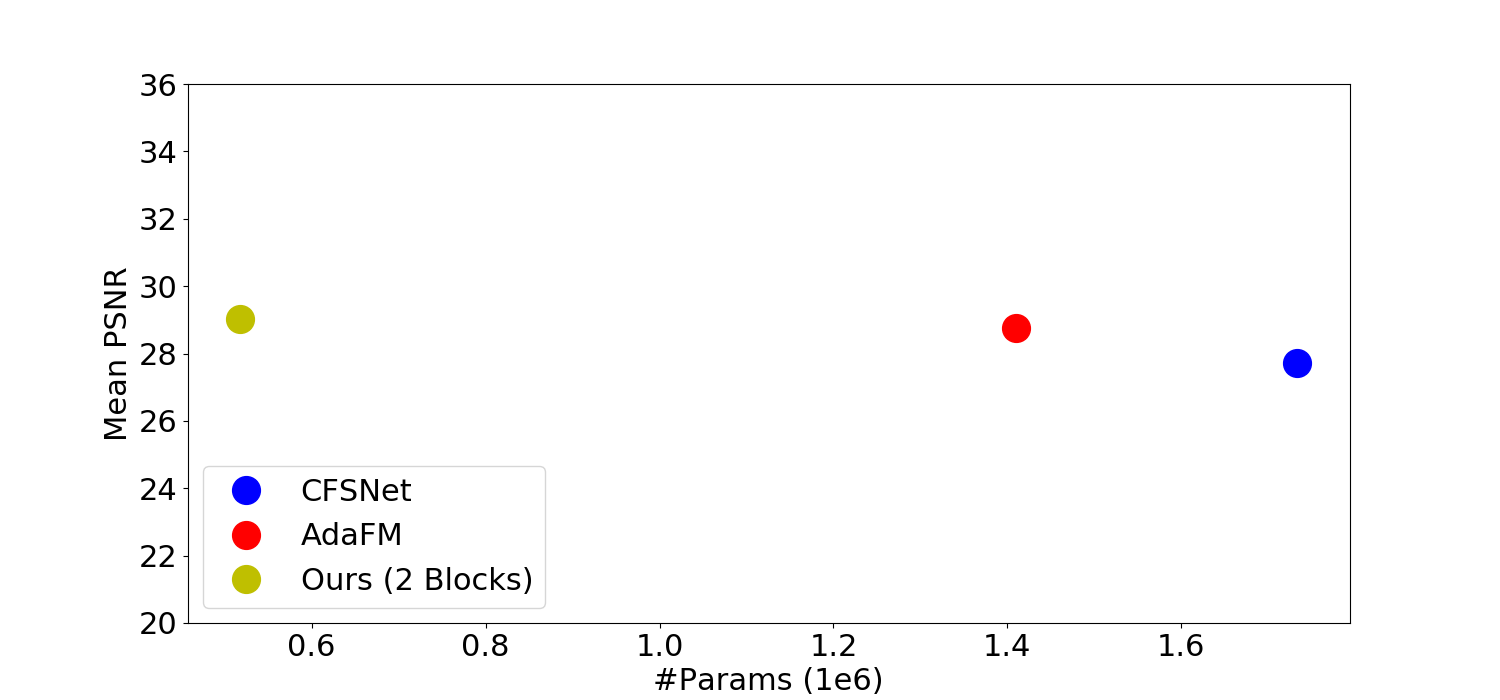}
	\end{center}
\caption{Results for super-resolution}
	\label{fig:weight_sr}
\end{figure}

Our evaluation is the same as before. Fig.~\ref{fig:weight_jpeg} shows our results for the DeJPEG task. Our model includes only 1 resblock with 0.37*1e6 parameters, AdaFM has 1.41*1e6 parameters, and CFSNet includes 1.73*1e6 parameters. Our model achieves slightly better accuracy with only 22\%-26\% of the parameters. For denoising (Fig.~\ref{fig:weight_noise}) our model achieves slightly better accuracy than other methods with a significant reduction of 44\%-54\% in the size of the network. For super-resolution (Fig.~\ref{fig:weight_sr}), our model yields comparable accuracy to AdaFM with only 36\% of the parameters and better accuracy than CFSNet with 29\% of the parameters. 

Overall, with respect to adaptive models that aim to reduce the number of parameters, our approach presents a significant saving in the number of parameters of up to 78\% with comparable accuracy. 


\subsection{Comparison with large size adaptive models}
 
We compare our results to those of the large-size adaptive model CResMD \cite{cresmd}. CResMD includes 32 residual blocks, 16 more than in our network. It has been shown that with these additional layers, CResMD outperforms other smaller-size adaptive methods.~\Cref{table:cres} presents the mean PSNR for both our approach and CResMD. Overall, our approach achieves better accuracy. 

\subsection{Input parameter tuning}

In the following, we explore the ability to accurately set the input parameters in our approach.

{ \bf Blind Setting}. We train a simple CNN (five convolutional layers and three fully connected layers) to estimate the degradation level of a noisy image. Based on our trained network, we can estimate the input degradation level and set the input scalar accordingly. On average, the degradation level estimation network achieves an accuracy of $98.23\%$. Overall, the estimation of noise level results in accurate restoration and can be advantageous in cases where the actual degradation level is unknown.

\begin{figure}[tb]
\centering
\includegraphics[width=0.75\linewidth]{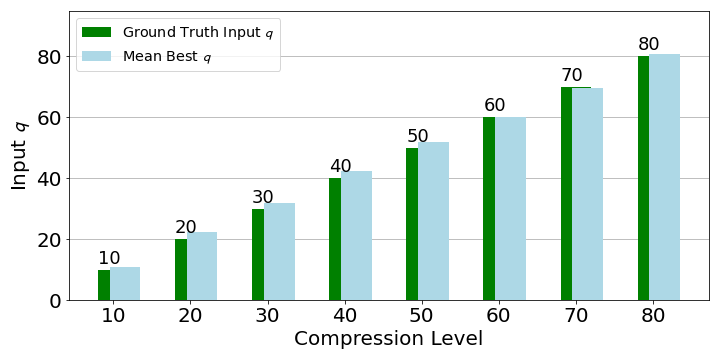}
    \caption{Parameter accuracy for DeJPEG. The dark green bars represent the ground truth levels while the light green bars represent the best input parameter that achieves the highest restoration accuracy.}
    \label{fig:linear_acc_a}
\end{figure}

{ \bf Manual Setting}. We experiment with the ability of the hypernetwork at inference time to generate, as trained, the optimal weights for the corresponding input degradation level. For each image in the test set, we degrade the image with a specific degradation level, e.g. $\sigma=15$ for the denoising task. For the degraded image, we measure the best input parameter that yields the network with the highest restoration accuracy in terms of PSNR and compare the value of the input parameter with the ground truth level.~\Cref{fig:linear_acc_a} presents the results, showing a negligible difference between the optimal and actual input.

\begin{table}[tb]
	\centering
	\footnotesize
	\caption{Mean PSNR}
	
	\begin{tabular}{lccc}
		  & \uline{DeJPEG} & \uline{denoise} & \uline{super-resolution}\\\cdashline{2-4} 
		CResMD & 32.9 & 28.80 & 28.71 \\ \cdashline{2-4}
		Ours   & 34.04 & 30.61 & 29.30 \\
		\end{tabular}

	\label{table:cres}
\end{table}

% \subsection{Implementation Details}

% The training data is augmented with both horizontal flipping and rotations. The mini-batch size is set to 16. We use randomly cropped $96 \times 96$ patches from each image as our training data. The $L1$ loss is used throughout all the experiments. We heuristically set the weight of each main network's loss uniformly. We use an initial learning rate set to $1\times 10^{-4}$, decaying by a factor of 10 after $5\times 10^5$ iterations. We train our model for $1\times 10^6$ iterations in total. The Adam optimizer is deployed with $\beta_1=0.9,\beta_2=0.999$. For super-resolution we train our model based on RGB images and evaluate the PSNR based on the y-channel. We down-scaled images by various factors and upscaled them to the original size using OpenCV bicubic interpolation (Pre-Upscaling SR). 
 
 \section{Conclusion}
We presented an efficient approach that can restore images with multiple degradation levels at runtime without the need to retrain the model. In order to increase efficiency, we propose using a hypernetwork-based model and simultaneously training several main networks. We demonstrate that our method achieves state-of-the-art accuracy while significantly reducing network size.

\vfill\pagebreak

\bibliographystyle{IEEEbib}
\bibliography{refs}

\end{document}